\pdfoutput=1
\documentclass[11pt]{article}

\usepackage[final]{coling}

\usepackage{times}
\usepackage{latexsym}

\usepackage[T1]{fontenc}
\usepackage[utf8]{inputenc}

\usepackage{microtype}

\usepackage{inconsolata}

\usepackage{graphicx}

\usepackage{multirow}
\usepackage{xcolor}
\usepackage{colortbl}
\usepackage{booktabs}
\usepackage{ulem}
\usepackage{array}
\usepackage{amsfonts,amssymb}
\usepackage{amsmath}
\usepackage[ruled,linesnumbered,noend]{algorithm2e}
\usepackage{algpseudocode}
\usepackage{wrapfig}
\usepackage{pifont}
\usepackage{arydshln}
\usepackage{marvosym}
\usepackage{enumerate}
\usepackage{enumitem}

\newcommand{\method}[1]{\textsc{#1}}
\newcommand{\model}{\method{HGCLIP}{}}

\title{\model: Exploring Vision-Language Models with Graph Representations for Hierarchical Understanding}

\author{Peng Xia$^{1,4}$, Xingtong Yu$^{3}$, Ming Hu$^1$, \textbf{Lie Ju$^1$,} \\ \textbf{Zhiyong Wang$^{2}$, Peibo Duan$^{1}$, Zongyuan Ge$^1$}\\ $^1$Monash University, $^2$ The University of Sydney,\\ $^3$  Singapore Management University, $^4$ UNC-Chapel Hill\\ \texttt{pxia@cs.unc.edu, zongyuan.ge@monash.edu}}

\begin{document}
\maketitle
\begin{abstract}
Object categories are typically organized into a multi-granularity taxonomic hierarchy. When classifying categories at different hierarchy levels, traditional uni-modal approaches focus primarily on image features, revealing limitations in complex scenarios. Recent studies integrating Vision-Language Models (VLMs) with class hierarchies have shown promise, yet they fall short of fully exploiting the hierarchical relationships. These efforts are constrained by their inability to perform effectively across varied granularity of categories.
To tackle this issue, we propose a novel framework (\textbf{\model}) that effectively combines \textbf{CLIP} with a deeper exploitation of the \textbf{H}ierarchical class structure via \textbf{G}raph representation learning. We explore constructing the class hierarchy into a graph, with its nodes representing the textual or image features of each category. After passing through a graph encoder, the textual features incorporate hierarchical structure information, while the image features emphasize class-aware features derived from prototypes through the attention mechanism. Our approach demonstrates significant improvements on 11 diverse visual recognition benchmarks. Our codes are fully available at \url{https://github.com/richard-peng-xia/HGCLIP}.
\end{abstract}

\section{Introduction}
\begin{figure}[t]
  % \vspace{-1em}
  \centering
  \begin{minipage}{0.49\linewidth}
    \centerline{\includegraphics[width=0.9\textwidth]{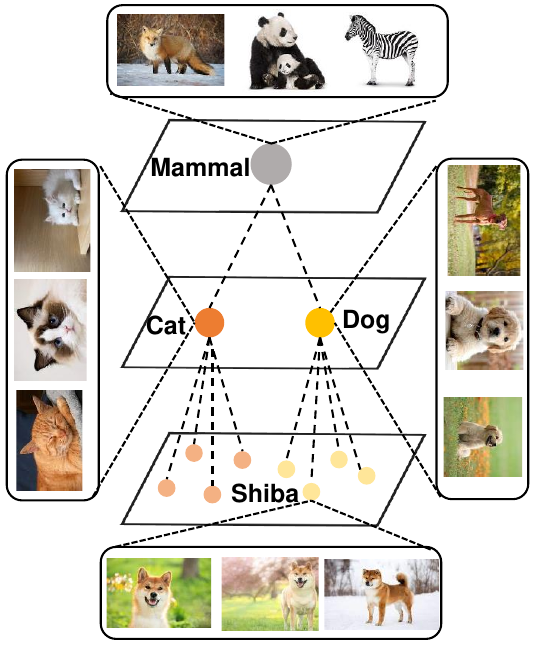}}
        \centerline{(a)}
        \label{fig:fig1a}
	\end{minipage}
    \label{fig:fig1b}
    \begin{minipage}{0.49\linewidth}
    \centerline{\includegraphics[width=0.9\textwidth]{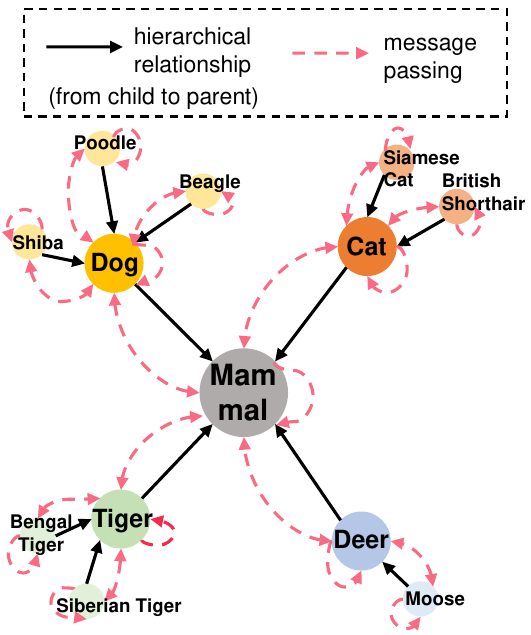}}
        \centerline{(b)}
        \label{fig:fig1b}
	\end{minipage}
 \vspace{-0.5em}
  \caption{An illustration of the graph representation based on class hierarchy. (a) The class hierarchy is presented in a tree structure. (b) The hierarchical labels are constructed into a graph, with nodes representing the text/image features of each class. The graph is fed into a graph encoder, where the nodes update the parameters by aggregating the messages from their neighboring nodes. Thus, the class features are fused with hierarchical information via graph representation learning.}
  \vspace{-1.5em}
  \label{fig:fig1}
\end{figure}

Hierarchical image classification~\cite{salakhutdinov2011learning,guo2018cnn} aims to enhance classification accuracy by identifying objects at various levels of granularity and capturing subtle relationships among them. Specifically, all the classes are organized into a multi-granularity taxonomic hierarchy (see in Figure~\ref{fig:fig1}a), where the top-level nodes represent broader categories (“Mammal''), while the lower-level nodes encompass finer-grained subcategories (“Dog''). The inherently hierarchical nature of the task compounds its complexity, as models must exhibit a keen understanding of semantic hierarchies, balancing the trade-off between capturing fine-grained details for subclasses while maintaining a broad understanding of superclasses~\cite{chen2018fine}. Previous works~\cite{chang2021your,guo2018cnn} mainly focus on enhancing image features according to the hierarchy of multiple branch outputs. These uni-modal methods only focus on the image modality, leading to certain limitations in complex scenarios, such as the inability to effectively utilize the textual descriptions of hierarchical labels and adapt to new classes or datasets. Therefore, leveraging multi-modal models (\textit{e.g.,} VLMs) to address hierarchical image classification presents stronger potential, offering richer information and greater scalability.

Given the powerful generalization capabilities of VLMs~\cite{radford2021learning,jia2021scaling,zhai2022lit} demonstrated on downstream tasks, harnessing their capabilities to address hierarchical image classification tasks presents a highly valuable exploration. These models are pre-trained on large-scale text-image pairs to align features from the image and text modalities in a shared latent embedding space. The predicted probabilities are obtained by calculating the similarity between image features and text features. \par Recently, some works have explored improving accuracy based on VLMs via class hierarchies. Specifically, CHiLS~\cite{novack2023chils} employs hierarchical mappings to transform each class into a list of subcategories. However, this approach has significant drawbacks when applied to fine-grained datasets, as the subcategories of these labels tend to be specialized and rare, resulting in an overly detailed and contextually sparse representation. Utilizing these specific labels as prompts may overwhelm the model, lacking broader contextual relevance. Hierarchy-CLIP~\cite{ge2023improving} proposes a label augmentation method that leverages the WordNet~\cite{fellbaum2010wordnet} label hierarchy, enriching each class with its parent and child classes. This method aims to provide a richer semantic expansion of class descriptions.
It enhances surface-level semantic associations rather than delving into the deeper and more structured connections inherent in a hierarchical structure. This limitation becomes apparent in scenarios requiring classification across multiple hierarchical levels, where a nuanced understanding of these relationships is crucial. Moreover, these methods are both training-free. While this offers the advantage of simplicity and direct application, it lacks the capacity for further model adaptation to specific datasets. Additionally, these methods do not fully exploit the potential of VLMs to adapt to the diverse and complex nature of hierarchical understanding.

Hence, the limitations of these approaches give rise to a new question: \textit{How can models leverage the class hierarchy thoroughly to simultaneously improve the prediction accuracy of categories at different semantic granularity levels?} 

To address this issue, we first introduce prompt learning~\cite{zhou2022learning,khattak2023maple} as an efficient method to adapt VLMs to downstream tasks. \model\ introduces prompt tokens within the multi-modal branches of CLIP to facilitate the learning of hierarchical contextual representations. More importantly, as demonstrated in Figure~\ref{fig:fig1}b, \model\ explores the integration of CLIP with graph representations for hierarchical image classification. Specifically, hierarchical relationships are modeled as a graph, given that they inherently form a tree-like structure. Based on this graph, we employ a graph encoder~\cite{velivckovic2017graph} to encode text features, enabling them to incorporate hierarchical structural information. Moreover, since image features represent features of individual patches/pixels rather than categories, we utilize prototype learning to represent image features for each category. Similarly, a graph encoder is leveraged to allow the prototypes to learn hierarchical relationships, and subsequently utilize the attention mechanism to enable the spatial feature map of images to focus more on the class-aware features derived from prototypes. On hierarchical image classification, \model\ outperforms existing CLIP-based approaches across both generic and fine-grained datasets. In scenarios where hierarchical labels are unavailable, \model\ also improves accuracy when utilizing class hierarchies queried by ChatGPT~\cite{openai2023chat}. Further, \model\ demonstrates favorable generalization ability and robustness in domain generalization and subpopulation shift settings, resulting in consistent improvements over existing methods.
To sum up, the main contributions of this work include: \vspace{-0.5em}
\begin{itemize}[leftmargin=*]
    \item We propose \model, a state-of-the-art (SoTA) method in hierarchical image classification for adaptation of CLIP. \vspace{-0.5em}
    \item To better utilize label hierarchies, we explore the graph representations to incorporate hierarchical structural information into vision-language feature representations for effective hierarchical understanding. \vspace{-0.5em}
    \item Our approach exhibits new SoTA performance across eleven hierarchical image classification benchmarks.
    \vspace{-0.5em}
\end{itemize}

\section{Related Work}
\textbf{Prompt Learning in Vision-Language Models:} 
VLMs leverage information from both image and text modalities to encode multimodal representations. VLMs,  e.g., CLIP~\cite{radford2021learning}, ALIGN~\cite{jia2021scaling}, and LiT~\cite{zhai2022lit} are pre-trained on large-scale image-text pairs and demonstrate remarkable representation abilities on various downstream tasks~\cite{gao2023clip,zhu2022prompt,ding2022decoupling,xia2024lmpt,xia2024rule,xia2024mmed,xia2024cares}. However, efficiently adapting them to downstream tasks is still a major challenge. Prompt learning~\cite{li2021prefix,lester2021power}, as a parameter-efficient technique, is well-suited for utilizing the representation capacity of pre-trained VLMs to boost performance, instead of the resource-intensive process of full fine-tuning. Many works~\cite{shu2022test,zhou2023zegclip,li2024tp} have demonstrated powerful performance on specific downstream tasks by combining VLMs and prompt tuning. \\
\textbf{Hierarchical Image Classification:} Hierarchical image classification~\cite{salakhutdinov2011learning,guo2018cnn} aims to categorize images into a hierarchical structure that organizes classes into a tree-like taxonomy. It acknowledges the inherent hierarchical nature of visual concepts, allowing for more nuanced and contextually rich image categorization. Prior research has explored various methodologies, including model architectures tailored for hierarchical classification~\cite{guo2018cnn,chang2021your,chen2018fine}, and exploiting the relationship of the categories in the hierarchy~\cite{liu2022focus}. Furthermore, the development of hierarchical classification has spurred some works~\cite{ju2023hierarchical,ju2024explore,yi2022exploring} that harness the class hierarchy across diverse domains, as these models tend to focus more on fine-grained and semantically relevant features. Recently, hierarchical labels are integrated with VLMs~\cite{novack2023chils,ge2023improving}. Nonetheless, these methods roughly overlook the hierarchical relationships among labels. Our work \textit{comprehensively} leverages the hierarchical relationships among labels, resulting in performance improvements on \textit{both generic and fine-grained} datasets.\\ 
\textbf{Graph Representation Learning:}\label{sec:grl}
Modern graph analysis methods rely on graph representation learning, encompassing graph embedding, graph neural networks (GNNs), and transformers. Early graph embedding techniques~\cite{perozzi2014deepwalk,grover2016node2vec} typically map nodes into a low-dimensional space, capturing structural information. Recently, GNNs~\cite{kipf2016semi,velivckovic2017graph} have become the mainstream technique in graph representation learning. They rely on a message-passing framework where each node refines its representation by recursively aggregating messages from its neighbors. Moreover, some recent approaches have also explored transformer-based architectures~\cite{yun2019graph,hu2020heterogeneous}. Furthermore, the boom of graph representation learning also advances the research and development in other communities such as CV \cite{shi2019skeleton} and NLP \cite{zhang2023pixel}. In this work, we employ \textit{hierarchical graph representations} to enrich multi-modal features, thus improving the model performance and generalization.

\begin{figure*}[t]
    \centering
    \includegraphics[width=0.95\linewidth]{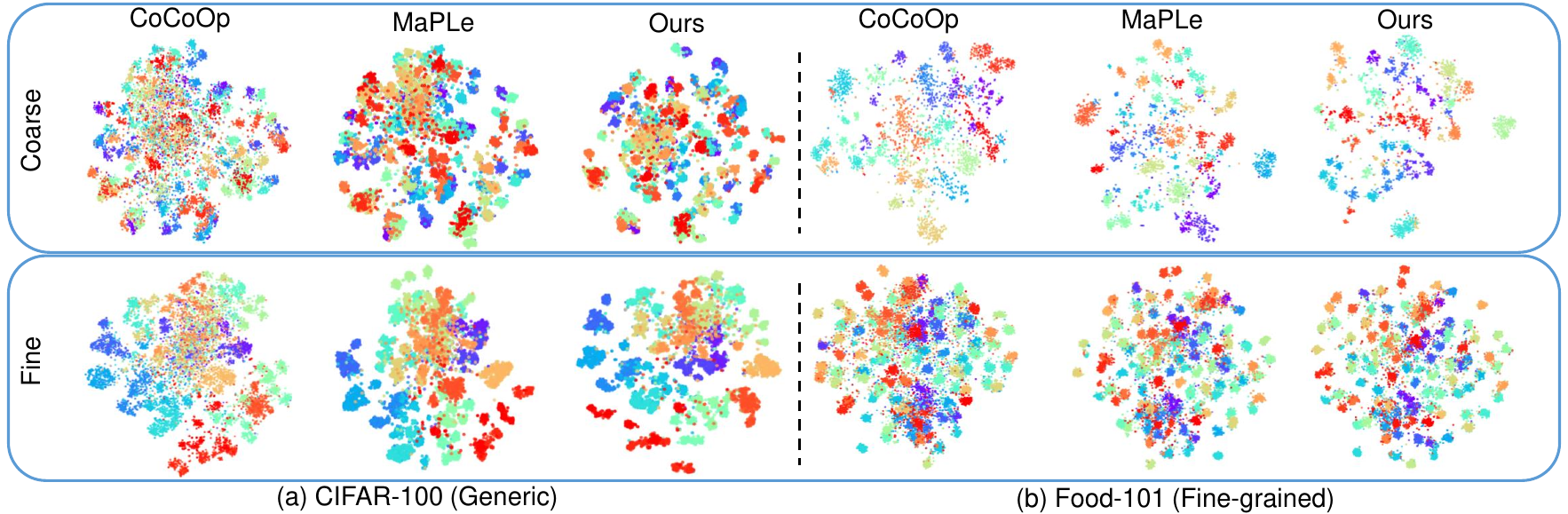}
    \vspace{-0.5em}
    \caption{t-SNE plots of image embeddings in SOTA method CoCoOp, MaPLe, and \model\ on two datasets with distinct semantic granularities. \model\ shows better separability in both fine-grained and coarse-grained levels.}
    \label{fig:tsne}
    \vspace{-1.5em}
\end{figure*}

\section{Preliminaries}
In this work, our goal is to learn hierarchical multi-modal knowledge via graph encoder based on CLIP. We will introduce related concepts and definitions in the following.
% $\boldsymbol{V}(\cdot)$ $\mathrm{ImageEnc}(\cdot)$ , both of which consist of $Q$ transformer layers $\{\mathcal{I}_{i}\}_{i=1}^Q \{\mathcal{T}_{i}\}_{i=1}^Q$
\subsection{Revisiting CLIP}
\quad We denote the CLIP image and text encoder as $\mathcal{I}(\cdot)$ and $\mathcal{T}(\cdot)$. The dataset contains $K$ categories, \textit{i.e.}, $\{C_1,\cdots,C_K\}$. CLIP leverages a structured approach by inserting all category names into a predefined textual template represented by the \texttt{[CLASS]} token, \textit{e.g.}, creating expressions like “\texttt{a photo of a [CLASS].}". This results in the generation of textual inputs denoted as $T_{K}$. Subsequently, textual features, represented as $\mathbf{F}_{t} \in \mathbb{R}^{K \times D}$, are extracted. Each input image $I$ is divided into $M$ fixed-sized patches, and each patch is embedded into $D$-dimensional latent space. Then CLIP derives its spatial feature map $\mathbf{F}_{s} \in \mathbb{R}^{H \times W \times D}$ and computes the global visual representations $\mathbf{f}_{v} \in \mathbb{R}^{1 \times D}$ through pooling operations, where $H$ and $W$ denote the height and width of the feature map. The integration of features from both encoders is achieved through cosine similarity measures, ultimately yielding classification $logits \in \mathbb{R}^{1 \times K}$. This comprehensive process can be summarized as follows
\begin{equation}
    \mathbf{F}_{t} = \mathcal{T}(T_{K}) ,
\end{equation}
\begin{equation}
    \mathbf{f}_{v} = \textsc{Pooling}(\mathbf{F}_{s}), \quad \textbf{F}_{s} = \mathcal{I}(I),
\end{equation}
\begin{equation}
    \textit{logits} = \mathbf{f}_{v}{\mathbf{F}_{t}}^{T}.
\end{equation}

The matrix multiplication operation between $\mathbf{f}_{v}$ and $\mathbf{F}_{t}$ is equivalent to calculating cosine similarities, assumed to be $L_2$-normalized features. $\textit{logits}$ signifies the computed probabilities for all $K$ categories, and CLIP identifies the category with the maximum output probability $\mathrm{argmax}_{C_K}(\textit{logits})$ as its final prediction.

\subsection{Graph Encoder}
\textbf{Graph.}
A graph is represented as \( G = (V, E) \), with \( V \) denoting the set of nodes and \( E \) the set of edges. 
Equivalently, the graph can be represented by an adjacency matrix $A$, such as $A_{ij}=1$, if $(v_i,v_j) \in E$, for any $v_i,v_j\in V$.\\
\textbf{Graph Encoder.}
GNNs are popular choices of graph encoder, most of which employ a message-passing mechanism \cite{wu2020comprehensive}. Specifically, each node in the graph aggregates messages (\textit{i.e.}, input features or embeddings) from its neighboring nodes to update its own embedding. Multiple layers of neighborhood aggregation can be stacked, facilitating recursive message passing across the graph. 
Formally, in the $l$-th GNN layer, the embedding of node $v$, denoted by $\mathbf{f}^l_v$, is calculated based on the embeddings in the previous layer, as follows
\begin{equation}
    \mathbf{f}^l_v = \textsc{Aggr}(\mathbf{f}^{l-1}_v, \{\mathbf{f}^{l-1}_u : u\in\mathcal{N}_v\}; \theta^l),
\end{equation}
where $\mathcal{N}_v$ is the set of neighboring nodes of $v$, $\theta^l$ is the learnable GNN parameters in layer $l$. $\textsc{Aggr}(\cdot)$ is the neighborhood aggregation function and can take various forms, ranging from the simple mean pooling \cite{kipf2016semi} to advanced neural networks such as neural attention \cite{velivckovic2017graph} or multi-layer perceptrons \cite{xu2018powerful}. Note that the initial node embedding $\mathbf{f}^0_v$ is simply given by the input feature.
We abstract the multi-layer encoding process as 
\begin{align}
    \mathbf{f}_v = \textsc{GraphEncoder}(\mathbf{f}_v^0,\mathcal{N}_v;\Theta),
\end{align}
where $\Theta=(\theta^1,\ldots,\theta^L)$ is the collection of weights across the layers. Note that graph embedding methods \cite{perozzi2014deepwalk,tang2015line,grover2016node2vec} and graph transformers \cite{yun2019graph,hu2020heterogeneous,ying2021transformers} could also serves as \textsc{GraphEncoder}. 

% \definecolor{customblue}{HTML}{9CC2DD}
% \definecolor{customyellow}{HTML}{FFE699}
\setlength{\fboxsep}{1pt}
\begin{figure*}[t]
    \centering
    \includegraphics[width=0.95\linewidth]{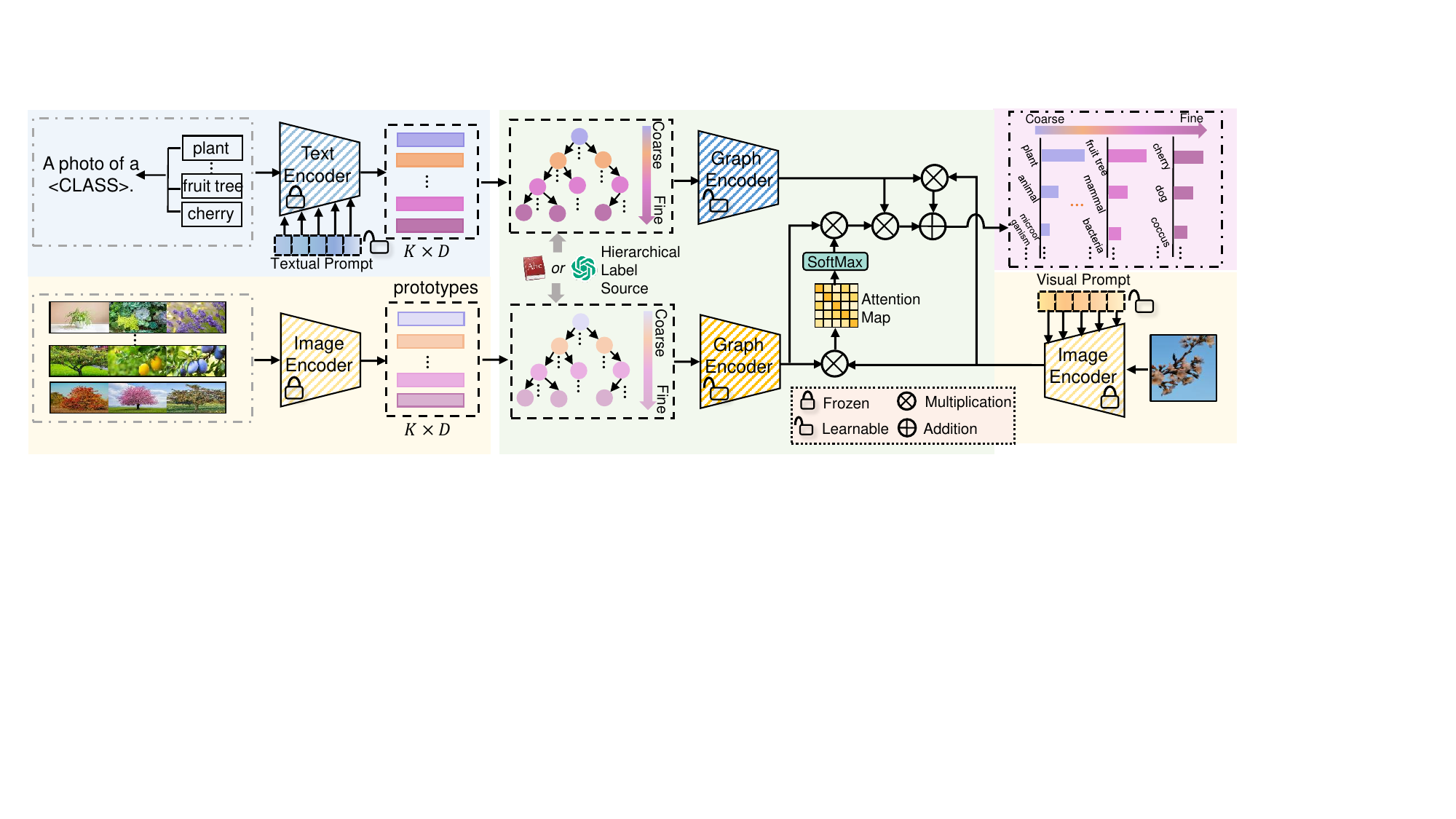}
    \caption{The pipeline of \model\ for adapting CLIP to hierarchical image classification. We introduce multi-modal hierarchical prompt to learn contextual representations. Then we construct the label hierarchy into a graph, with its nodes representing the textual or image features of each class. Features integrate hierarchical structure information through message passing in the graph encoder. Textual features directly combine hierarchical representations, while image features focus on class-aware prototypes through the attention mechanism.}
    \label{fig:framework}
    \vspace{-1.5em}
\end{figure*}

\section{Methodology}
\quad In this section, as shown in Figure~\ref{fig:framework}, we present our proposed method, \textit{i.e.}, \model\, for adapting pre-trained VLMs for hierarchical understanding. Our approach aims to enhance the capacity for understanding multiple semantic levels. Most prior approaches focus on single-label classification, whereas hierarchical classification necessitates that the model attends to features relevant to multi-granularity hierarchies. To this end, \model\ entails: \textbf{a)} introducing learnable prompt tokens within multiple transformer blocks in both the visual and textual branches to learn hierarchical contextual representations; \textbf{b)} employing a graph encoder to encode textual features, integrating them with hierarchical structural information; \textbf{c)} utilizing prototype learning to represent image features of each category and similarly modeling them utilizing a graph encoder, thereafter employing the attention mechanism to enable the spatial feature map of images to focus more on class-aware and hierarchy-guided image features.

\subsection{Hierarchy Setting}
\quad The ground truth class hierarchy currently available in a dataset is usually obtained by querying a WordNet~\cite{fellbaum2010wordnet}-like dictionary, but in the real world, our dataset may have no available class hierarchy. In this case, we turn to LLMs, \textit{i.e.}, ChatGPT, to approximate the hierarchy diagram. Specifically, given some label set size $K$, semantic granularity levels $h$, class names, and optional context, we query ChatGPT with the prompt: \par \texttt{Generate $h$-tier hierarchical labels for the following $K$ categories: $\{C_1,\cdots,C_K\}$.}

\subsection{Multi-modal Hierarchical Prompt}
\quad In order to comprehensively and efficiently leverage the capabilities of pretrained VLMs, we explore the potential of multi-modal prompt, encompassing both textual and visual prompt. As highlighted in~\cite{khattak2023maple}, the acquisition of prompt at deeper transformer layers is crucial, as it progressively models hierarchical feature representations. Learnable tokens are introduced at multiple transformer blocks of both textual and visual branches of VLMs, given as textual prompt $\mathbf{P}^T=\{\mathbf{p}^T_1,\cdots,\mathbf{p}^T_t\}$ and visual prompt $\mathbf{P}^V=\{\mathbf{p}^V_1,\cdots,\mathbf{p}^V_v\}$, respectively. Therefore, the image encoder processes the input tokens added visual prompt $\mathbf{P}^V$ to generate prompted spatial feature map represented as $\tilde{\mathbf{F}_{s}} \in \mathbb{R}^{(HW +v)\times D}$ and prompted global visual representations $\tilde{\mathbf{f}_{v}} \in \mathbb{R}^{1 \times D}$. Similarly, textual prompt $\mathbf{P}^T$ are incorporated into the input tokens for encoding, and textual features are obtained as $\tilde{\mathbf{F}_{t}} \in \mathbb{R}^{K \times D}$. These hierarchical prompt tokens leverage the knowledge encoding capabilities of VLMs to effectively learn task-relevant contextual representations across different semantic levels. 

\subsection{Delving into Graph Representations}
\quad The hierarchical structure among labels naturally forms a tree structure, hence we leverage graph representations to model the hierarchy and integrate it into multi-modal features. In Figure~\ref{fig:tsne}, we visualize and compare the image embeddings of \model\ with those of previous SoTA CoCoOp and MaPLe. It is worth noting that the image embeddings of CLIP, CoOp, CoCoOp, and KgCoOp would be identical, as they do not learn prompts in the visual branch. The visualization reveals that the image embeddings of \model\ are more separable, indicating that incorporating hierarchical information can better adapt CLIP. \\
\textbf{Encoding Text:}
Clearly, textual features $\tilde{\mathbf{F}_{t}}=\{\tilde{\mathbf{f}_{n}^{t}}\}_{n=1}^K$ can be directly employed as input for a graph encoder, as they possess corresponding $D$-dimensional textual features for each category. 
The class hierarchy is constructed into a graph, where vertices and edges represent individual classes and pairs of classes with hierarchical relationships, respectively. As a result, each node $n$ of the text-attributed graph is associated with text features of the corresponding category $\tilde{\mathbf{f}_{n}^{t}}$. The graph encoder approaches node classification by using the structural interactions between nodes. The textual features $\hat{\mathbf{F}_{t}}=\{\hat{\mathbf{f}_{n}^{t}}\}_{n=1}^K$ integrating hierarchical information are encoded as follows:
\begin{equation}
    \{\hat{\mathbf{f}_{n}^{t}}\}_{n=1}^K = \textsc{GraphEncoder}(\tilde{\mathbf{f}_{n}^{t}},\mathcal{N}_n;\Theta_t),
\end{equation}
where $\Theta_t$ denotes the parameters of the graph encoder for textual modality, $\mathcal{N}_n$ denotes the neighbor nodes of $n$. \\
\textbf{Encoding Image:}
In contrast to textual features, the spatial feature map represents the features of each patch, and the global visual representations characterize the image holistically, rather than representing features for each category. Therefore, the image features of each image cannot be directly input into the graph encoder.
\begin{figure}
    \centering
    % \vspace{-1em}
    \includegraphics[width=0.4\textwidth]{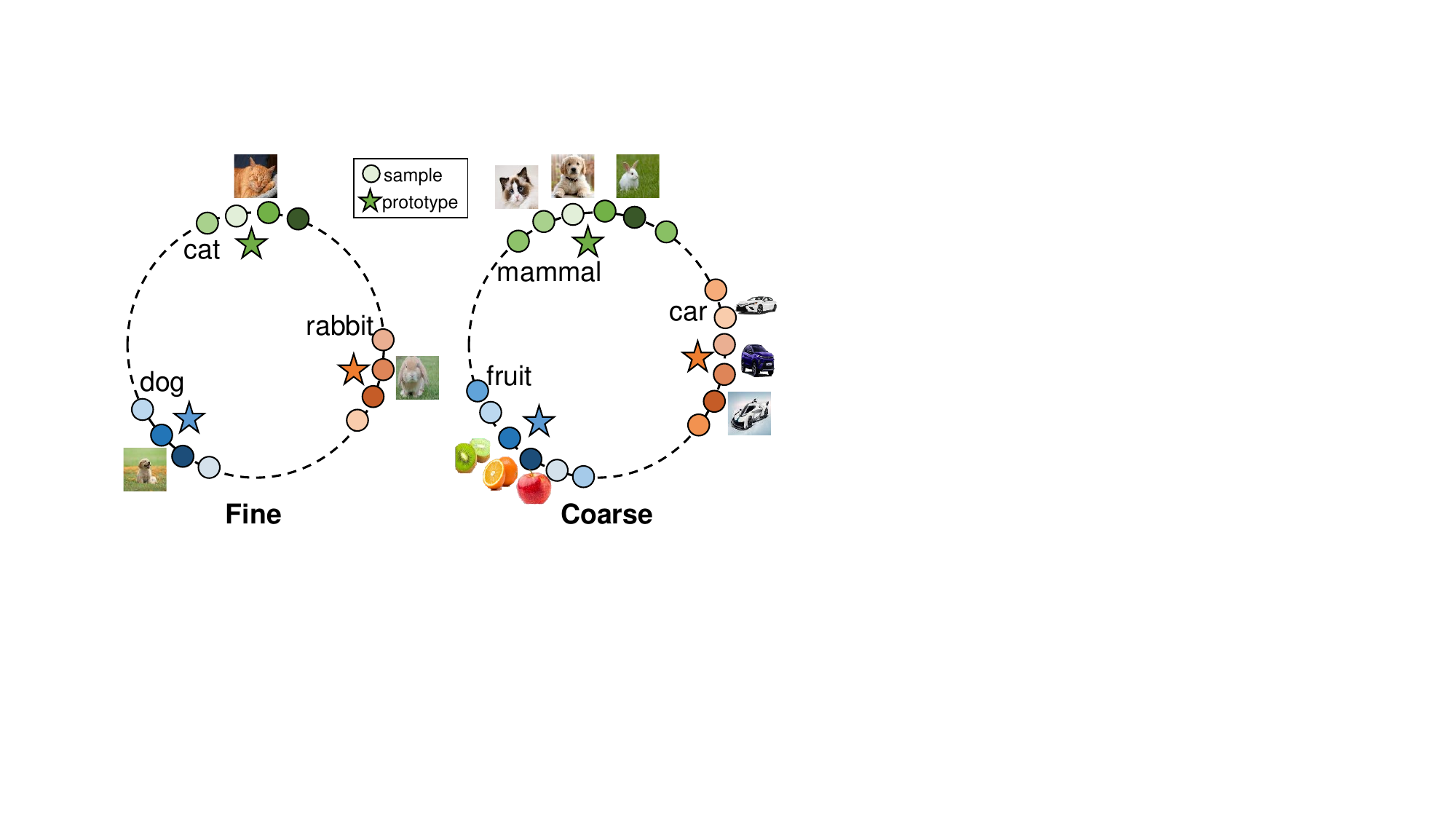}
    \caption{Semantic prototypes are constructed to guide the learning of hierarchical semantics of images.}
    \vspace{-1em}
    \label{fig:prototype}
\end{figure}
% \mathbf{F}_{s}^{*}\in \mathbb{R}^{K \times M\times M \times D}
To address this issue, as shown in Figure~\ref{fig:prototype}, we first leverage prototype learning to represent the image features for each category. The global features $\mathbf{F}_{v}^{*}=\{\mathbf{f}_{n}^{v*}\}_{n=1}^K \in \mathbb{R}^{K \times 1 \times D}$ of all images $\{I_{n}\}_{n=1}^K$ (only in the training set) belonging to each class are extracted as prototypes for all categories. These prototypes can then be utilized by the graph encoder to be encoded. The procedure is as follows
\begin{equation}
    \mathbf{F}_{v}^{*} = \textsc{Pooling}(\mathbf{F}_{s}^{*}),\quad \mathbf{F}_{s}^{*} = \mathcal{I}(I_{K}).
\end{equation}

In the image-attributed graph, each node $n$ is associated with image features $\{\mathbf{f}_{n}^{v*}\}_{n=1}^K$, while the rest is consistent with the text-attributed graph. Similarly, the image features $\hat{\mathbf{F}_{v*}}$ are encoded as follows
\begin{align}
   \small \{\hat{\mathbf{f}_{v*}^{n}}\}_{n=1}^K = \textsc{GraphEncoder}(\mathbf{f}_{v*},\mathcal{N}_n;\Theta_v),
\end{align}
% \vspace{-0.5em}

where $\Theta_v$ denotes the parameters of the graph encoder for visual modality. After the visual graph encoder effectively leverages structural knowledge, we then employ the attention mechanism to obtain the attention weights of visual features $\tilde{\mathbf{F}_{s}}$ with respect to the prototypes $\hat{\mathbf{F}_{v*}}$. The calculation of attention weights is as follows
\begin{equation}
    \psi = \tilde{\mathbf{F}_{s}}{{\hat{\mathbf{F}_{v*}}}^T} \in \mathbb{R}^{(HW+v) \times K},
\end{equation}

where $\psi$ denotes the attention map. Each element of $\psi$ represents the attention weight, namely, the feature similarity between a class prototype and one image pixel/site. Based on $\psi$, we update the spatial feature map as follows
\begin{equation}
     \hat{\mathbf{F}_{s}} = \mathrm{SoftMax}(\psi/\alpha)\hat{\mathbf{F}_{v*}},
\end{equation}
where $\alpha$ modulates the attention magnitude. Weighted by the attention scores representing similarity, the image features incorporate structured information from the prototypes. As the prototypes $\hat{\mathbf{F}_{v*}}$ encode $K$-category visual knowledge, the signals of classes appearing in the image would be more notable. Meanwhile, the spatial feature map provides pixel-level fine-grained information for the interaction, contributing to the thorough integration of class-aware features from the prototypes into the image features. \\
\textbf{Classification Logits:} Finally, we obtain the attention-interacted global visual feature by pooling and output the classification logits as
% \begin{figure}[t]
%     \centering
%     \includegraphics[width=0.5\linewidth]{img/prototypes_.pdf}
%     \caption{Semantic prototypes are constructed to guide the learning of hierarchical semantics of images.}
%     \label{fig:prototype}
%     % \vspace{-1.5em}
% \end{figure}
\begin{equation}
    \hat{\mathbf{f}_{v}} = \textsc{Pooling}(\hat{\mathbf{F}_{s}}) \in \mathbb{R}^{1 \times D},
\end{equation}
\begin{equation}
    \textit{logits} = \lambda_1 \cdot \tilde{\mathbf{f}_{v}}{\hat{\mathbf{F}_{t}}}^{T}+\lambda_2 \cdot \hat{\mathbf{f}_{v}}{\hat{\mathbf{F}_{t}}}^{T},
\end{equation}
where $\lambda_1$ and $\lambda_2$ denote hyper-parameters to control the weight assigned to the $\textit{logits}$ that incorporate structured image features.

For hierarchical image classification, the model is required to simultaneously predict several labels at different granularities. Consequently, it is necessary to partition the predicted $\textit{logits}$ into their respective hierarchical categories $\textit{logits}_i$, with each level corresponding to the ground truth labels $GT_i$, where $i=1,\cdots,h$. The overall loss function can be defined as follows
\begin{equation}
    \mathcal{L} = \sum_{i=1}^{h}w_i\cdot\mathcal{L}_{CE}({GT}_i, \textit{logits}_i),
\end{equation}
where $w_i$ denotes the weights for learning features at different hierarchical levels and $\mathcal{L}_{CE}(\cdot,\cdot)$ represents a cross-entropy loss. A higher $w_i$ prioritizes the learning of features at the $i$-th level, and vice versa.

\begin{table*}[t]
\centering
\scriptsize
\resizebox{\linewidth}{!}{
\begin{tabular}{lccccccccc}
\toprule \rowcolor{gray!20}
\multicolumn{2}{l|}{Dataset}  & \begin{tabular}[c]{@{}c@{}}CLIP \\ \textcolor{gray}{\scriptsize \textup{ICML'21}}\end{tabular}  & \begin{tabular}[c]{@{}c@{}}CoOp \\ \textcolor{gray}{\scriptsize \textup{IJCV'22}}\end{tabular} &  \begin{tabular}[c]{@{}c@{}}CoCoOp \\ \textcolor{gray}{\scriptsize \textup{CVPR'22}}\end{tabular} &  \begin{tabular}[c]{@{}c@{}}VPT \\ \textcolor{gray}{\scriptsize \textup{ECCV'22}}\end{tabular} & \begin{tabular}[c]{@{}c@{}}MaPLe \\ \textcolor{gray}{\scriptsize \textup{CVPR'23}}\end{tabular} & \begin{tabular}[c]{@{}c@{}}KgCoOp \\ \textcolor{gray}{\scriptsize \textup{CVPR'23}}\end{tabular} & \begin{tabular}[c]{@{}c@{}}PromptSRC \\ \textcolor{gray}{\scriptsize \textup{ICCV'23}}\end{tabular} & \begin{tabular}[c]{@{}c@{}}\model\\ \textcolor{gray}{\scriptsize \textup{(Ours)}}\end{tabular} \\ \hline \hline 
\multirow{2}{*}{CIFAR-100*}   & \multicolumn{1}{c|}{$l_1$} & 43.22  & 83.76  & 82.60  & 88.75  & \underline{90.67} & 78.65  & 88.18  & \textbf{91.87}  \\
& \multicolumn{1}{c|}{$l_2$} & 66.57  & 76.81  & 75.73  & 83.94  & \underline{85.81} & 73.49  & 82.24  & \textbf{86.55}  \\ \hline
\multirow{3}{*}{Caltech-101} & \multicolumn{1}{c|}{$l_1$} & 58.47  & 96.19  & 96.95 & 96.62  & \underline{98.06}  & 93.50  & 95.70  & \textbf{98.50}   \\
& \multicolumn{1}{c|}{$l_2$} & 69.01  & 95.12  & 94.18 & 95.06  & \underline{97.38}  & 93.56  & 95.57  & \textbf{97.51}   \\
& \multicolumn{1}{c|}{$l_3$} & 84.56  & 95.88  & 95.85 & 96.12  & \underline{96.88}  & 94.81  & 95.51  & \textbf{97.03}   \\ \hline
\multirow{3}{*}{\begin{tabular}[l]{@{}l@{}}FGVC-\\Aircraft*\end{tabular}} & \multicolumn{1}{c|}{$l_1$} & 31.08  & 54.30  & 54.80 & 56.81  & \underline{70.79}  & 53.31  & 55.99  & \textbf{79.24}   \\
& \multicolumn{1}{c|}{$l_2$} & 35.49  & 51.59  & 50.42  & 53.00  & \underline{68.87} & 50.38  & 50.20  & \textbf{70.70}   \\
& \multicolumn{1}{c|}{$l_3$} & 24.69  & 37.74  & 36.10  & 35.00 & \underline{52.58} & 35.56  & 34.21  & \textbf{61.33}   \\ \hline
\multirow{2}{*}{\begin{tabular}[l]{@{}l@{}}Stanford\\Cars*\end{tabular}} & \multicolumn{1}{c|}{$l_1$} & 61.75  & 82.39  & 83.31  & \underline{83.46}  & 83.35 & 82.78 & 82.85  & \textbf{83.61}  \\
& \multicolumn{1}{c|}{$l_2$} & 63.59  & 73.35  & 73.86  & 76.53  & \underline{76.92} & 69.01 & 73.24  & \textbf{77.84}  \\ \hline
\multirow{2}{*}{Food-101} & \multicolumn{1}{c|}{$l_1$} & 61.38 & 87.04 & 89.13 & \underline{89.28} & 88.16 & 85.18 & 87.37 & \textbf{91.12} \\
& \multicolumn{1}{c|}{$l_2$} & 85.53  & 86.94 & 87.10  & \underline{88.46}  & 88.04 & 86.00 & 86.89 & \textbf{88.73}  \\ \hline
\multirow{3}{*}{Fruits-360} & \multicolumn{1}{c|}{$l_1$} & 75.43  & 93.73  & 99.21 & 99.46  & \underline{99.59} & 90.95  & 94.41  & \textbf{99.71}   \\
& \multicolumn{1}{c|}{$l_2$} & 34.40  & 86.87  & 96.46 & \textbf{98.65}  & 98.31 & 76.43  & 94.60  & \underline{98.51}   \\
& \multicolumn{1}{c|}{$l_3$} & 23.55  & 85.90  & 96.58 & 97.12  & \textbf{97.78} & 72.80  & 92.13  & \underline{97.73}   \\  \hline   
\multirow{2}{*}{\begin{tabular}[l]{@{}l@{}}Oxford\\Pets-37\end{tabular}} & \multicolumn{1}{c|}{$l_1$} & \textbf{99.97}  & 99.86  & 99.86 & \underline{99.92}  & 99.91 & 99.89  & 99.89  & \underline{99.92}   \\
& \multicolumn{1}{c|}{$l_2$} & 88.14  & 91.94  & 91.97 & 91.81 & 92.14 & 91.81  & \underline{92.19}  & \textbf{92.31}   \\ \hline
\multirow{2}{*}{EuroSAT} & \multicolumn{1}{c|}{$l_1$} & 62.97 & 91.50 & 91.77 & 92.41 & \underline{93.03} & 91.38 & 91.96 & \textbf{95.57} \\
& \multicolumn{1}{c|}{$l_2$} & 41.01 & 86.68 & 87.36 & 88.94 & \underline{90.16} & 87.88 & 88.30 & \textbf{92.79}  \\ \hline
\multirow{3}{*}{SUN397*} & \multicolumn{1}{c|}{$l_1$} & 70.29 & 88.01 & 88.60 & \underline{90.82} & 90.59 & 87.62 & 88.18 & \textbf{92.16}   \\
& \multicolumn{1}{c|}{$l_2$} & 63.59 & 84.28 & 84.20 & 86.29 & \underline{86.72} & 85.11 & 85.40 & \textbf{88.39}  \\
& \multicolumn{1}{c|}{$l_3$} & 60.85 & 78.67 & 78.66 & 80.66 & \underline{81.37} & 79.16 & 80.08 & \textbf{83.41} \\  \hline   
\multirow{2}{*}{DTD} & \multicolumn{1}{c|}{$l_1$} & 55.17 & 80.86 & 80.98 & \underline{83.83} & 83.41 & 87.45 & 81.83 & \textbf{86.82}  \\
& \multicolumn{1}{c|}{$l_2$} & 48.09 & 74.60 & 72.19 & 78.34 & \underline{78.36} & 75.68 & 75.79 & \textbf{81.08}  \\ \hline
\multirow{4}{*}{ETHEC*} & \multicolumn{1}{c|}{$l_1$} & 31.12  & 89.45  & 89.61 & 90.87 & \underline{92.17} & 86.03 & 90.64 & \textbf{95.76}   \\
& \multicolumn{1}{c|}{$l_2$} & 2.65  & 85.10  & 86.07 & 86.79  & \underline{89.60} & 83.11  & 87.91  & \textbf{93.40} \\ & \multicolumn{1}{c|}{$l_3$} & 17.94  & 74.67  & 75.02 & 75.81  & \underline{78.48} & 71.33 & 77.46  & \textbf{82.98} \\ & \multicolumn{1}{c|}{$l_4$} & 1.52 & 49.48  & 51.27 & 51.99  & 55.73  & 47.62  & \underline{55.75} & \textbf{60.39}  \\ \bottomrule
\end{tabular}
}
\vspace{-1em}
\caption{Top-1 accuracy (\%) comparison on hierarchical image classification of \model\ with previous CLIP-based prompt tuning methods. The best result is \textbf{bold} and the second best is \underline{underlined}. * denotes that the dataset is with available class hierarchy, and hierarchies of others are queried through ChatGPT. $l_i$ represents the classification accuracy at the $i$-th hierarchical level, where a smaller $i$ indicates a coarser granularity level, and vice versa.}
\label{tab:hie}
\vspace{-1em}
\end{table*}

\section{Experiment}
\subsection{Benchmark Setting}
\textbf{Hierarchical Image Classification:} We consider 11 visual classification datasets, covering a wide range of recognition tasks. These include two general object datasets, CIFAR-100~\cite{krizhevsky2009learning} and Caltech-101~\cite{fei2004learning}; six fine-grained datasets, FGVC-Aircraft~\cite{maji2013fine}, StanfordCars~\cite{krause20133d}, Food-101~\cite{bossard2014food}, Fruits-360~\cite{murecsan2017fruit}, OxfordPets-37~\cite{parkhi2012cats} and ETHEC~\cite{dhall2020hierarchical}; a scene recognition dataset SUN397 \cite{xiao2010sun}; a texture dataset DTD \cite{cimpoi2014describing} and a satellite image dataset EuroSAT \cite{helber2019eurosat}. The aim is to demonstrate our method under general situations of data diversity, where the label hierarchical levels range from two to four. \\
\textbf{Implementation Details:} We use top-1 accuracy to evaluate the prediction performance. We adopt CLIP ViT-B/16 as the visual encoder and use the corresponding CLIP Transformer as the text encoder. We set $\lambda_1$ = 1 and $\lambda_2$ = 0.2 to weight the proportion of hierarchical structural information. For hierarchical classification, we use deep prompting with $v=t=4$ in the first 9 transformer layers and train the models for 50 epochs. All models are trained with a batch size of 64 and a learning rate of 3e-4 via SGD optimizer, and decay by the cosine annealing rule during training. 

\subsection{Hierarchical Image Classification}
\textbf{CLIP-based prompt tuning methods.} Table~\ref{tab:hie} displays the comparative performance of zero-shot CLIP, recent works on prompt learning and \model\ on 11 diverse hierarchical classification datasets. In the case of CLIP, we utilize handcrafted specific prompts designed for each dataset. In comparison with state-of-the-art MaPLe~\cite{khattak2023maple} and PromptSRC~\cite{khattak2023self}, \model\ exhibits improved performance across all levels on all the datasets, with the exception of a slight decline in performance on Fruits-360. With the contribution of graph representations, as opposed to SoTA MaPLe and PromptSRC, \model\ demonstrates superior generalization across multiple hierarchical categories on all the datasets, achieving an absolute average gain of 2.2\% and 5.7\% respectively. \\
\noindent \textbf{CLIP-based feature adaptation methods.} In Table~\ref{tab:feature}, we compare \model\ with prior feature adaption methods based on CLIP. CLIP-Adapter~\cite{gao2023clip} learns two residual-style adapters after CLIP. Tip-Adapter~\cite{zhang2022tip} constructs a key-value cache model by extracting features from few-shot data, then views the cache model as a well-performing initialization and fine-tunes the cache keys. CALIP~\cite{guo2023calip} is proposed to boost CLIP performance via a parameter-free attention module between multi-modal representations. In comparison with these feature adaption approaches, \model\ exhibits excellent feature representation capabilities, with an accuracy on CIFAR-100 that is 8.7\%, 6.2\%, and 13.3\% higher than theirs, respectively. \\
\noindent \textbf{Visual-only hierarchical image classification methods.} We have analysed various multi-modal methods above, and to demonstrate the effectiveness of \model, we compare visual-only fine-grained visual classification methods, as shown in Table~\ref{tab:fine}. Our method still achieve a significant advantage. Additionally, the visual-only FGVC methods are more time-consuming compared to ours (100 v.s. 50 training epochs). 

\vspace{-0.5em}
\begin{table}[h]
    \centering
    \footnotesize
    \resizebox{\linewidth}{!}{
    \begin{tabular}{lcc}
    \toprule
    \multirow{2}{*}{Method}          & \multicolumn{2}{c}{Acc. \%} \\ \cmidrule(r){2-2} \cmidrule(r){3-3}          & $l_1$      & $l_2$      \\
    \midrule
    CLIP~\cite{radford2021learning}~\textcolor{gray}{\scriptsize \textup{ICML'21}} & 43.22      &  66.57    \\
    Linear probe  & 75.60 & 71.27 \\
    CLIP-Adapter~\cite{gao2023clip}~\textcolor{gray}{\scriptsize \textup{IJCV'23}}  & 83.91     & 77.03     \\
    Tip-Adapter~\cite{zhang2022tip}~\textcolor{gray}{\scriptsize \textup{ECCV'22}} & 84.57     & 81.42     \\ 
    CALIP~\cite{guo2023calip}~\textcolor{gray}{\scriptsize \textup{AAAI'23}} & 77.51 & 74.28 \\ \rowcolor{gray!20}
    \model~\textcolor{gray}{\scriptsize \textup{(Ours)}} & \textbf{91.87} & \textbf{86.55} \\ \bottomrule
    \end{tabular}
    }
    \vspace{-1em}
    \caption{Comparision with CLIP-based feature adaption methods.}
    \label{tab:feature}
    \vspace{-1em}
\end{table}

\begin{table}[htbp]
    \centering
    \footnotesize
    \begin{tabular}{lcc}
    \toprule
    \multirow{2}{*}{Method}  & \multicolumn{2}{c}{Acc. \%} \\ \cmidrule(r){2-2} \cmidrule(r){3-3} & $l_1$  & $l_2$ \\
    \midrule
    PMG~\cite{du2020fine}~\textcolor{gray}{\tiny \textup{ECCV'20}}  & 87.16 & 83.02 \\
    FGN~\cite{chang2021your}~\textcolor{gray}{\tiny \textup{CVPR'21}}  & 87.88 & 83.60 \\
    GHORD~\cite{zhao2021graph}~\textcolor{gray}{\tiny \textup{CVPR'21}}  & 87.93 & 84.36 \\
    CHRF~\cite{liu2022focus}~\textcolor{gray}{\tiny \textup{ECCV'22}} & 88.67 & 84.91 \\ 
    TFGIC~\cite{xu2023trusted}~\textcolor{gray}{\tiny \textup{AAAI'23}} & 89.20 & 85.17 \\ \rowcolor{gray!20}
    \model~\textcolor{gray}{\tiny \textup{(Ours)}} & \textbf{91.87} & \textbf{86.55}  \\ \bottomrule
    \end{tabular}
    \vspace{-1em}
    \caption{Comparison with visual-only SOTA FGVC methods.}
    \label{tab:fine}
    \vspace{-1em}
\end{table}

\begin{table}[htbp]
    \centering
    \scriptsize
    \resizebox{\linewidth}{!}{
    \begin{tabular}{ccccccccc}
    \toprule
    \multicolumn{4}{c}{Module} & \multicolumn{2}{c}{CIFAR-100} & \multicolumn{3}{c}{FGVC-Aircraft} \\ \cmidrule(r){5-6}  \cmidrule(r){1-4} \cmidrule(r){7-9}
    \multicolumn{1}{c}{TP}  & \multicolumn{1}{c}{TG}  & \multicolumn{1}{c}{VP}  & \multicolumn{1}{c}{VG} & \multicolumn{1}{c}{$l_1$} & \multicolumn{1}{c}{$l_2$} & \multicolumn{1}{c}{$l_1$} & \multicolumn{1}{c}{$l_2$} & \multicolumn{1}{c}{$l_3$}\\ \midrule
    \ding{55} & \ding{55} & \ding{55} & \ding{55} & 43.22  & 66.57 & 31.08  & 35.49 & 24.69 \\
    \ding{51} & \ding{55} & \ding{55} & \ding{55} & 84.21  & 77.22 & 54.96  & 52.67 & 38.72 \\
    \ding{55} & \ding{55} & \ding{51} & \ding{55} & 84.21  & 77.22 & 56.81  & 53.00 & 35.00 \\
    \ding{51} & \ding{51} & \ding{55} & \ding{55} & 87.42  & 81.24 & 61.56  & 57.90 & 42.83 \\
    \ding{55} & \ding{55} & \ding{51} & \ding{51} & 87.18  & 80.87 & 61.52  & 58.13 & 43.17 \\
    \ding{51} & \ding{55} & \ding{51} & \ding{55} & 90.67  & 85.81 & 70.79  & 68.87 & 52.58 \\
    \ding{51} & \ding{55} & \ding{51} & \ding{51} & 91.28 & 86.04 & 74.61  & 69.27 & 55.66 \\
    \ding{51} & \ding{51} & \ding{51} & \ding{55} & 91.43 & 85.96 & 75.37  & 69.28 & 57.50 \\ \rowcolor{gray!20}
    \ding{51} & \ding{51} & \ding{51} & \ding{51} & \textbf{91.87} &  \textbf{86.55} & \textbf{79.24}  & \textbf{70.70} & \textbf{61.33} \\ \bottomrule 
    \end{tabular}
    }
    \vspace{-1em}
    \caption{Component Analysis of \model. TP and VP serve as textual and visual prompts. TG and VG denote graph encoder for textual and visual modality.}
    \label{tab:aba}
    \vspace{-1em}
\end{table}

\subsection{Ablative Analysis}
\textbf{Components Analysis}. \model\ primarily consists of multi-modal prompts and graph encoders. In Table~\ref{tab:aba}, we ablate on the performance of each module. The prompts facilitate the model in learning hierarchical contextual features, while the graph encoders effectively integrate hierarchical structure information into the feature representations. This enables the model to achieve impressive results across multiple semantic granularities. \\
\noindent \textbf{Noisy Hierarchies Queried by LLMs}. It is important to note that LLMs may output sub-optimal hierarchical labels. LLMs produce inconsistent hierarchical labels based on a set of input category names or generate hierarchical labels of different levels, leading to certain biases in the model performance. However, even when utilizing the noisy hierarchy, \model\ still enhances accuracy within the original categories in the dataset. 

\begin{figure}[htbp]
    \centering
    \includegraphics[width=0.45\textwidth]{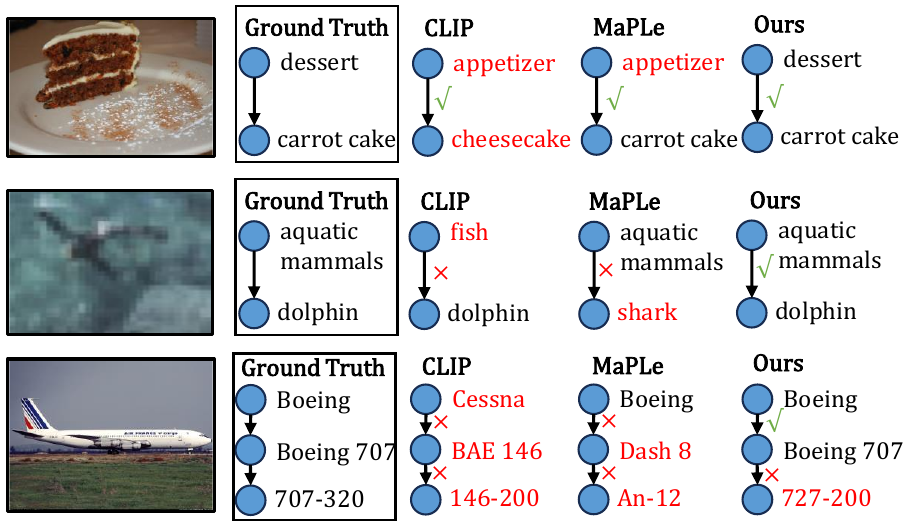}
    \caption{Example decisions from our model, MaPLe and CLIP.}
    \label{fig:case}
    \vspace{-1.5em}
\end{figure}

\subsection{Qualitative Analysis}
Figure~\ref{fig:case} presents illustrative cases showcasing the predicted probabilities of the models at different semantic granularity levels. CLIP shows inconsistencies in classification results at different levels, indicating that CLIP does not grasp the semantic relationship between different hierarchical levels. MaPLe improves prediction accuracy via learning hierarchical feature representation. However, it still displays inconsistencies when predicting classifications across different levels. Our method largely mitigates this issue, leveraging hierarchical graph representation to bolster the learning of inter-level class features.

\subsection{Graph Encoder Analysis}
\begin{figure}[htbp]
    \centering
    \includegraphics[width=0.85\linewidth]{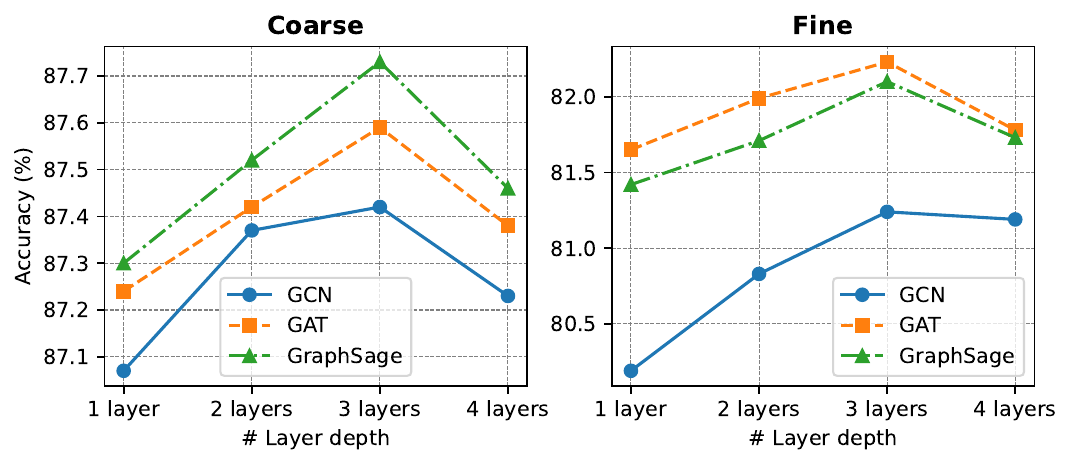}
    \caption{The impact of various GNN variants and the number of layers on hierarchical classification. We report accuracy results on two hierarchical levels.}
    \label{fig:graph}
    \vspace{-1.5em}
\end{figure}

We further conduct experiments to analyze the impact of various graph encoders. We apply three of the most commonly used graph learning models: GCN~\cite{kipf2016semi}, GAT~\cite{velivckovic2018graph}, and GraphSAGE~\cite{hamilton2017inductive} to \model, the results are illustrated in Figure~\ref{fig:graph}. First, for both hierarchical levels, GAT consistently exhibits superior performance, particularly at the fine-grained level, where GAT surpasses the other encoders. Second, with the increase of layer depth of the graph encoder, the accuracy initially rises. Upon reaching a peak (3 layers), the accuracy begins to gradually decline with further increase in layer depth. Therefore, we use a 3-layer GAT as graph encoder in our experiments.

\section{Conclusion}
In this work, we propose a novel view that combines VLMs with graph representations. Deep prompts are incorporated into the multi-modal branches, allowing VLMs to better learn hierarchical representations. The graph-based hierarchical relationships are encoded into features, strengthening the connection of features across multiple granularities. When integrating image features with graph representation, given that image features are pixel/region-level, prototype learning is employed for class-level image features, which are then fused with the image features through the attention mechanism. \model\ achieves SoTA results on several hierarchical image classification benchmarks.

\newpage
\section*{Limitations} Although this work has achieved results by utilizing graph representations to characterize hierarchical information, it merely employs a simple graph encoder (such as GCN, GAT, GraphSAGE) to encode structural information. With the advancement of graph learning, it is anticipated that there will be better graph learning methods for representing hierarchical structures, which could further enhance features. We separately employ graph encoders for different modalities, yielding the best performance, outperforming cross-modal graph encoders with shared weights. This may be because current multi-modal models lack the ability to accurately extract features from each modality. With the development of multimodal models, this limitation is expected to be addressed, significantly reducing training time and improving inference speed. Additionally, we leverage multi-modal prompt learning but have not integrated it with graph learning, perhaps combining the two could yield better results.

\bibliography{custom}

\appendix

\section{Data Details}
\label{sec:appendix}

\subsection{Hierarchical Classes Number} Table~\ref{tab:num} presents the number of categories at each hierarchical level for all the datasets utilized in the experiments.

\begin{table}[htbp]
\centering
\footnotesize
\begin{tabular}{lcc}
\toprule 
\multicolumn{2}{l|}{Dataset}  & Classes  \\ \hline
\multirow{2}{*}{CIFAR-100~\cite{krizhevsky2009learning}}   & \multicolumn{1}{c|}{$l_1$} & 20  \\
& \multicolumn{1}{c|}{$l_2$} & 100  \\ \hdashline[1pt/1pt]
\multirow{3}{*}{Caltech-101~\cite{fei2004learning}} & \multicolumn{1}{c|}{$l_1$} & 11$^\dag$   \\
& \multicolumn{1}{c|}{$l_2$} & 51$^\dag$   \\
& \multicolumn{1}{c|}{$l_3$} & 101   \\ \hdashline[1pt/1pt]
\multirow{3}{*}{FGVC-Aircraft~\cite{maji2013fine}} & \multicolumn{1}{c|}{$l_1$} & 30   \\
& \multicolumn{1}{c|}{$l_2$} & 70   \\
& \multicolumn{1}{c|}{$l_3$} & 100   \\ \hdashline[1pt/1pt]
\multirow{2}{*}{StanfordCars~\cite{krause20133d}} & \multicolumn{1}{c|}{$l_1$} & 9  \\
& \multicolumn{1}{c|}{$l_2$} & 196  \\ \hdashline[1pt/1pt]
\multirow{2}{*}{Food-101~\cite{bossard2014food}} & \multicolumn{1}{c|}{$l_1$} & 18$^\dag$ \\
& \multicolumn{1}{c|}{$l_2$} & 101  \\ \hdashline[1pt/1pt]
\multirow{3}{*}{Fruits-360~\cite{murecsan2017fruit}} & \multicolumn{1}{c|}{$l_1$} & 4$^\dag$ \\
& \multicolumn{1}{c|}{$l_2$} & 69$^\dag$   \\
& \multicolumn{1}{c|}{$l_3$} & 113   \\  \hdashline[1pt/1pt] 
\multirow{2}{*}{OxfordPets-37~\cite{parkhi2012cats}} & \multicolumn{1}{c|}{$l_1$} & 2$^\dag$  \\
& \multicolumn{1}{c|}{$l_2$} & 37   \\ \hdashline[1pt/1pt]
\multirow{3}{*}{SUN397~\cite{xiao2010sun}} & \multicolumn{1}{c|}{$l_1$} & 3  \\
& \multicolumn{1}{c|}{$l_2$} & 15 \\
& \multicolumn{1}{c|}{$l_3$} & 397 \\ \hdashline[1pt/1pt]
\multirow{2}{*}{DTD~\cite{cimpoi2014describing}} & \multicolumn{1}{c|}{$l_1$} & 5$^\dag$  \\
& \multicolumn{1}{c|}{$l_2$} & 10   \\ \hdashline[1pt/1pt]
\multirow{2}{*}{EuroSAT~\cite{helber2019eurosat}} & \multicolumn{1}{c|}{$l_1$} & 9$^\dag$  \\
& \multicolumn{1}{c|}{$l_2$} & 47   \\ \hdashline[1pt/1pt]
\multirow{4}{*}{ETHEC~\cite{dhall2020hierarchical}} & \multicolumn{1}{c|}{$l_1$} & 6   \\
& \multicolumn{1}{c|}{$l_2$} & 21 \\ 
& \multicolumn{1}{c|}{$l_3$} & 135 \\ 
& \multicolumn{1}{c|}{$l_4$} & 561  \\ \hline
\multirow{2}{*}{ImageNet*~\cite{deng2009imagenet}} & \multicolumn{1}{c|}{$l_1$} & 328  \\
& \multicolumn{1}{c|}{$l_2$} & 314   \\ \hline
\multirow{2}{*}{Living17~\cite{santurkar2021breeds}} & \multicolumn{1}{c|}{$l_1$} & 17  \\
& \multicolumn{1}{c|}{$l_2$} & 34   \\ \hdashline[1pt/1pt]
\multirow{2}{*}{Nonliving26~\cite{santurkar2021breeds}} & \multicolumn{1}{c|}{$l_1$} & 26 \\
& \multicolumn{1}{c|}{$l_2$} & 52   \\ \hdashline[1pt/1pt]
\multirow{2}{*}{Entity13~\cite{santurkar2021breeds}} & \multicolumn{1}{c|}{$l_1$} & 13  \\
& \multicolumn{1}{c|}{$l_2$} & 130   \\ \hdashline[1pt/1pt]
\multirow{2}{*}{Entity30~\cite{santurkar2021breeds}} & \multicolumn{1}{c|}{$l_1$} & 30  \\
& \multicolumn{1}{c|}{$l_2$} & 120   \\
\bottomrule
\end{tabular}
\caption{The number of categories at each hierarchical level for all datasets. $\dag$ indicates that the categories for that level are queried through ChatGPT.}
\label{tab:num}
% \vspace{-1.5em}
\end{table}

\subsection{Textual Diversity} Table~\ref{tab:prompt} displays the hand-crafted prompts used for textual prompt initialization. 

\begin{table}[htbp]
\centering
\footnotesize
\resizebox{\linewidth}{!}{
\begin{tabular}{lc}
\toprule 
Dataset & Hand-crafted prompt  \\ \hline
CIFAR-100  & \texttt{a photo of a [CLASS].}  \\
Caltech-101 & \texttt{a photo of a [CLASS].}  \\
FGVC-Aircraft & \texttt{a photo of an aircraft [CLASS].}  \\
StanfordCars & \texttt{a photo of a car [CLASS].}  \\
Food-101 & \texttt{a photo of a food [CLASS].}  \\
Fruits-360 & \texttt{a photo of fruits [CLASS].}  \\
OxfordPets-37 & \texttt{a photo of pets [CLASS].}  \\
SUN397 & \texttt{a photo of [CLASS].}  \\
DTD  & \texttt{a photo of texture [CLASS].}  \\
EuroSAT & \texttt{a centered satellite photo of [CLASS].}  \\
ETHEC & \texttt{a photo of a butterfly [CLASS].} \\ \hline
ImageNet & \texttt{a photo of a [CLASS].}  \\
ImageNetV2 & \texttt{a photo of a [CLASS].} \\
ImageNet-Sketch & \texttt{a photo of a [CLASS].} \\
ImageNet-A & \texttt{a photo of a [CLASS].} \\
ImageNet-R & \texttt{a photo of a [CLASS].} \\ \hline
Living17 & \texttt{a photo of a [CLASS].} \\
Nonliving26 & \texttt{a photo of a [CLASS].} \\
Entity13 & \texttt{a photo of a [CLASS].} \\
Entity30 & \texttt{a photo of a [CLASS].} \\ 
\bottomrule
\end{tabular}
}
\caption{Hand-crafted prompt for different datasets.}
\label{tab:prompt}
% \vspace{-1.5em}
\end{table}

\subsection{Dataset Setting}
\textbf{Fruits-360}~\cite{murecsan2017fruit}: From the perspective of class names, its original form is somewhat incomplete, as some classes are distinguished only by numerical indices (\textit{e.g.}, “Apple Golden 1" and “Apple Golden 2"). Therefore, we manually rename the classes, which leads to a reduction in the number of subclasses from 131 to 113. This makes the evaluation of model performance more reasonable and accurate. \newline
\textbf{BREEDS}~\cite{santurkar2021breeds}: Since our method necessitates the use of class hierarchies, we opt for the \textit{'good'} split of subpopulations within a given superclass into source and target domains, signifying that subpopulations in the source and target domains share a common parent. In contrast, the other splits \textit{i.e.,} \textit{'bad'} and \textit{'random'} would result in a greater degree of separation between the subpopulations of the source and target domains in terms of hierarchical structure distance. However, such an arrangement would not effectively validate the efficacy of our method. Although these two ways of partitioning the subpopulations would lead to stronger adversarial conditions, it would cause the model to learn the wrong label hierarchy. Thus, they are not considered in our analysis. 

\section{Implementation Details} 
\label{sup:a3}
We use a publically available ViT-B/16 CLIP model with $D$ = 512 and use a learning rate of 3e-4 which is fixed for all experiments in all benchmarks. We set $\lambda_1$ = 1 and $\lambda_2$ = 0.2 to weight the proportion of hierarchical structural information. The corresponding hyperparameters are fixed across all datasets and benchmark. The respective epochs are fixed across all datasets. For hierarchical classification, we use deep prompting with $v=t=4$ in the first 9 transformer layers and train for 50 epochs. Since the classification of coarse-grained categories is noticeably less challenging than that of fine-grained ones, we set $w_h$ in Eq. 13 to 2 and the remaining $\{w_i\}_{i=1}^{h-1}$ to 1, in order to improve the model performance on fine-grained classification. For distribution shift benchmarks, we use the same number of prompt tokens in the first 3 layers and train for 20 epochs. All models are trained using SGD optimizer and utilize 4 NVIDIA GeForce RTX 4090 GPUs.

\section{Distribution Shifts}
\begin{figure}[htbp]
    \centering
    \includegraphics[width=0.4\textwidth]{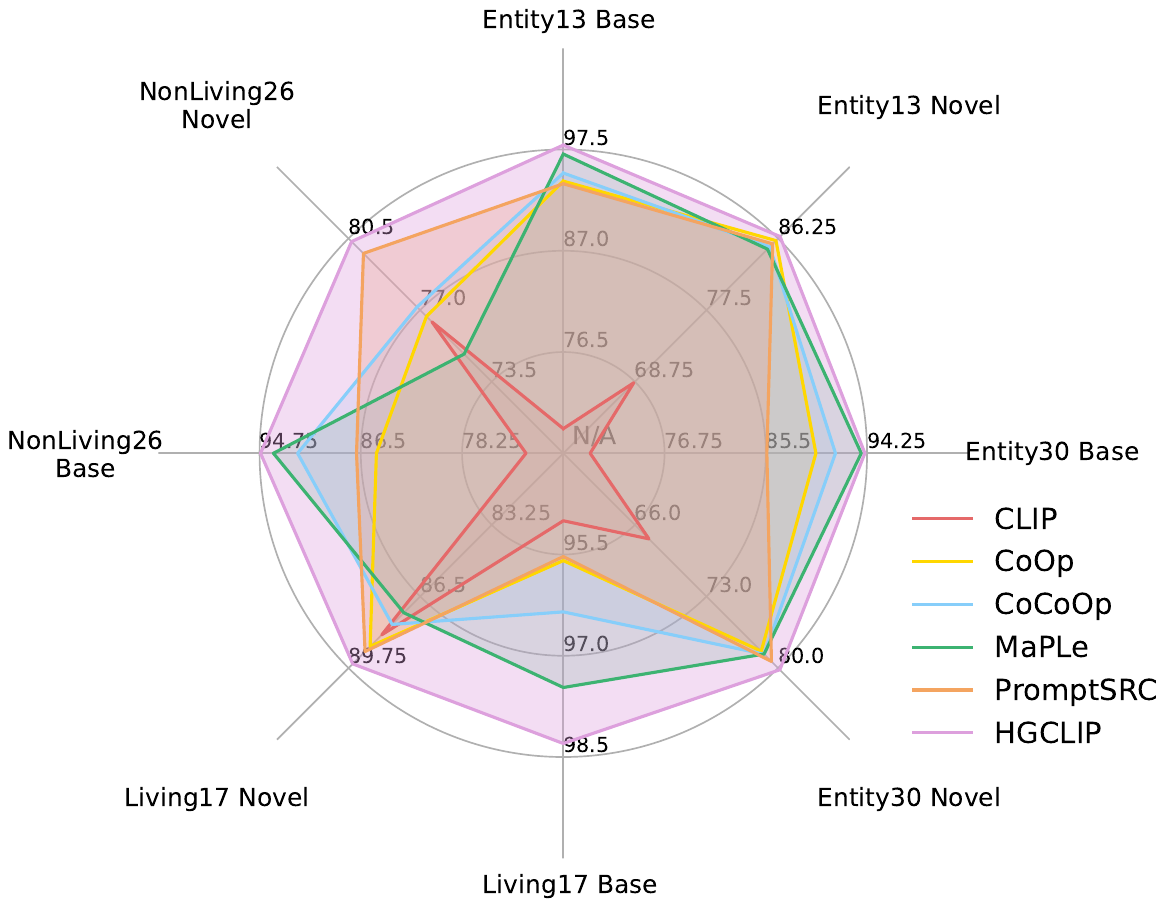}
    \caption{Results for different methods on the four BREEDS datasets~\cite{santurkar2021breeds} to measure robustness to subpopulation shift.}
    \label{fig:breeds}
\end{figure}

To validate the generalization and robustness of our proposed \model, we perform experiments on the two benchmarks with distributional shifts, namely domain shifts and subpopulation shifts. This allows us to comprehensively assess the efficacy of \model\ in generalizing to out-of-distribution datasets. \\
\newline
\textbf{Domain Generalization:}
We evaluate the robustness of our approach on out-of-distribution datasets~\cite{xia2024generalizing,hu2024diffusion}. The source distributions correspond to the original ImageNet~\cite{deng2009imagenet}. The task is to classify images from the target datasets (ImageNetV2~\cite{recht2019imagenet}, ImageNet-Sketch~\cite{wang2019learning}, ImageNet-A~\cite{hendrycks2021natural} and ImageNet-R~\cite{hendrycks2021many}), which consist of images that contain various types of domain shifts. It is important to note that due to the inconsistent semantic granularity at each hierarchical level in the original ImageNet, we only select all the categories from ImageNet-A and ImageNet-R to facilitate our experiment. Table~\ref{tab:domain} summarizes the results of \model\ and prior approaches on out-of-distribution datasets. We verify the transferability of models trained on ImageNet to various ImageNet variants with domain shifts. On the target datasets, the performance of \model\ surpasses previous SoTA methods. This achievement underscores the efficacy of hierarchical graph representations to enhance multi-modal features. Such an integration improves the generalization capabilities, enabling it to perform well across varying domains. This indicates that \model\ not only captures the intricate relationships within the data but also adapts effectively to new and unseen domains. \\
\noindent \textbf{Subpopulation Shift:}
We also examine \model's robustness to subpopulation shift within a dataset. The source and target distributions, though encompassing identical class categories, feature distinct subpopulations within those classes. Our empirical investigations were executed on the four BREEDS~\cite{santurkar2021breeds} ImageNet subsets: Living17, Nonliving26, Entity13, and Entity30. We further evaluate the generalizability of \model\ on four datasets from BREEDS (subsets of ImageNet) that exhibit subpopulation shifts and provide available class hierarchies. Figure~\ref{fig:breeds} depicts the performance of \model\ and previous methods under subpopulation shifts. The models are trained only on base classes and tested on novel classes. The results suggest that \model\ possesses strong generalization capabilities even when confronted with feature-level shifts, underscoring the efficacy of hierarchical structure information in enhancing model generalizability. This success demonstrates the robustness of \model\ in handling variations within subpopulations, ensuring consistent accuracy and reliability across diverse and shifting data landscapes.

\begin{table}[htbp]
    \centering
    \scriptsize
    \resizebox{\linewidth}{!}{
    \begin{tabular}{lcccccc}
    \toprule
    & \multicolumn{1}{c}{\textbf{Source}} & \multicolumn{4}{c}{\textbf{Target}} \\ \cmidrule(r){2-2}  \cmidrule(r){3-6}
    \multirow{-3}{*}{Method} &  \multicolumn{1}{c}{ImageNet} & \multicolumn{1}{c}{-V2}  & \multicolumn{1}{c}{-Sketch}  & \multicolumn{1}{c}{-A}  & \multicolumn{1}{c}{-R} \\ \midrule
    CLIP & 86.11 & 80.19 & 72.13 & 46.11 & 70.85 \\
    CoOp & 85.44 & 79.71 & 70.85 & 45.63  & 70.91 \\
    CoCoOp & 85.03 & 79.49 & 72.96 & 46.99 & 72.47 \\
    MaPLe  & \underline{91.22} & \underline{85.12} & 74.19 & 49.28 & 73.79 \\
    PromptSRC & 89.89  & 83.79 & \underline{75.04} & \underline{49.37} & \underline{74.22}  \\ \rowcolor{gray!20}
    \model\ & \textbf{92.19} & \textbf{86.24} & \textbf{77.40} & \textbf{50.38} & \textbf{76.07} \\ \bottomrule
    \end{tabular}
    }
    % \parbox{0.93\linewidth}{\scriptsize * denotes that 314 categories out of 1,000 categories are selected for the dataset.}
    \caption{Domain generalization. Comparison with CLIP-based methods on robustness to domain shifts. }
    \label{tab:domain}
\end{table}

\section{Method Introduction}
\subsection{Calculating Logits for Hierarchical Classification}
\begin{figure}[t]
    \centering
    \includegraphics[width=0.85\linewidth]{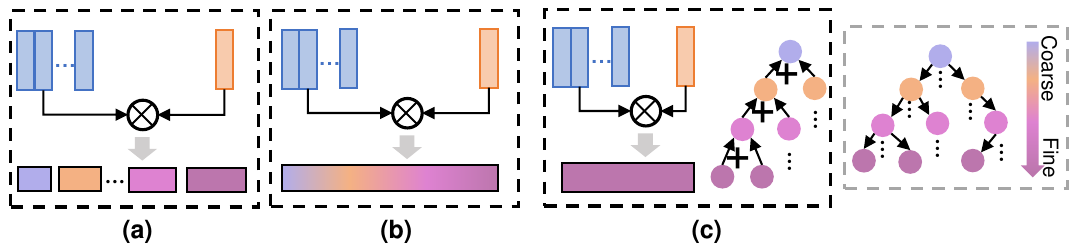}
    \caption{(a) Model schematic for the per-level classifier. (b) Model schematic for the hierarchy-agnostic classifier. (c) Model schematic for the marginalization method.}
    \label{fig:logits}
\end{figure}

The calculation of the predictive probabilities for each hierarchical level primarily involves three methods~\cite{dhall2020hierarchical}. The first method, as depicted in Figure~\ref{fig:logits}(a), describes a classifier that functions on a per-level basis, designated as $h$ individual $\{K_i\}_{i=1}^h$-way classifiers. The model utilizes information about the label hierarchy by explicitly predicting a single label per level for a given image. However, this computational approach cannot be applied to methods based on CLIP. This is because VLMs determine the predicted class by calculating the similarity between the features of the image and those of each text prompt, and the model does not possess multiple linear layers to correspond to the prediction results of each hierarchical level separately. Second, Figure~\ref{fig:logits}(b) illustrates a model configured as a hierarchy-agnostic classifier. This particular multi-label classifier operates without incorporating any knowledge of an explicit hierarchical structure within the labels. Our experiments are based on this computational method to calculate the predictive probabilities at each hierarchical level. Third, Figure~\ref{fig:logits}(c) presents a model schematic for the marginalization approach, wherein, rather than assigning a discrete label to each level, the model generates a probabilistic distribution across the terminal nodes of the hierarchical structure. The probability associated with non-terminal nodes is computed by marginalizing across their immediate offspring. This marginalization method not only captures the connections of the various nodes within the hierarchy but also acknowledges the presence of $h$ levels within the label hierarchy. However, employing this method to compute the predictive probabilities of superclasses yields poor results. Detailed comparative results are presented in Table~\ref{tab:logit}.
\newline
\begin{figure}[htbp]
    \centering
    \includegraphics[width=0.9\linewidth]{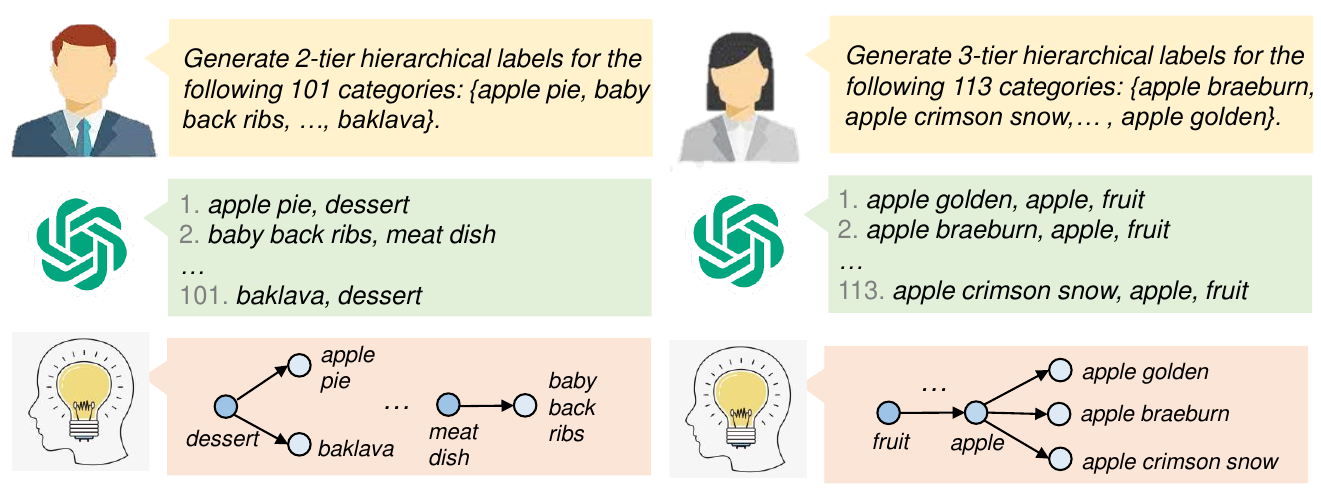}
    \caption{\centering Illustrations of GPT-generated hierarchical labels.}
    \label{fig:gpt}
    \vspace{-1.5em}
\end{figure}

\subsection{GPT-mapped Hierarchical Labels}
This is an illustrative diagram of hierarchical labels generated by ChatGPT \cite{openai2023chat}. As demostrated in Figure \ref{fig:gpt}, the prompts include all original class names and the desired number of hierarchy levels. ChatGPT will output the specified hierarchical labels.

\begin{table*}[htbp]
\footnotesize
\centering
\resizebox{\linewidth}{!}{
\begin{tabular}{lccccccc}
\toprule
\multicolumn{1}{c}{\multirow{2}{*}{Method}} & \multirow{2}{*}{Prompt} & \multicolumn{2}{c}{CIFAR-100} & \multicolumn{4}{c}{ETHEC} \\ \cmidrule(r){3-4} \cmidrule(r){5-8}
\multicolumn{1}{c}{}  &  & $l_1$ & $l_2$ & $l_1$ & $l_2$ & $l_3$ & $l_4$  \\ \hline
\multirow{3}{*}{Marginalization}  & \multirow{1}{*}{\texttt{a photo of a [CLASS](, a type of butterfly).}} & 6.13  & 66.61 & 10.76 & 1.65 & 0.1 & 1.5 \\ 
  & \multirow{1}{*}{80 Prompt Templates Ensemble} & 6.35  & 66.97 & 10.76 & 1.65 & 0.1 & 1.31  \\
  & \multirow{1}{*}{\texttt{a photo of a [CLASS$_{l_2}$], which is a kind of [CLASS$_{l_1}$].}} & 5.57  & 67.99 & 10.76 & 1.65 & 0.1 & 1.42  \\ \hline
\multirow{4}{*}{Multi-Label}  & \multirow{1}{*}{\texttt{a photo of a [CLASS](, a type of butterfly).}} & 43.22  & 66.57 & \textbf{31.12} & 2.65 & \textbf{17.94} & \textbf{1.52}  \\ 
  & \multirow{1}{*}{80 Prompt Templates Ensemble} & 44.85  & 66.97 & 20.82 & 2.29 & 16.58 & 1.29  \\
\multirow{2}{*}{}  & \texttt{a photo of a [CLASS$_{l_1}$], possibly a [CLASS$_{l_2}$].} & \textbf{57.83}  & 66.57  & 27.58 & \textbf{11.60} & 17.88 & \textbf{1.52}    \\
& \multicolumn{1}{l}{\texttt{a photo of a [CLASS$_{l_2}$], which is a kind of [CLASS$_{l_1}$].}}  & 43.22  & \textbf{68.00} & \textbf{31.12} & 1.65 & 0.1 & 1.42  \\ 
\bottomrule     
\end{tabular}}
\caption{Comparative results of the probability calculations for categories arcoss different hierarchical levels using multiple prompt templates.}
\label{tab:logit}
\end{table*}

\section{Ablative Analysis}
\label{sup:c}
\begin{figure}[htbp]
    \centering
    \includegraphics[width=0.6\linewidth]{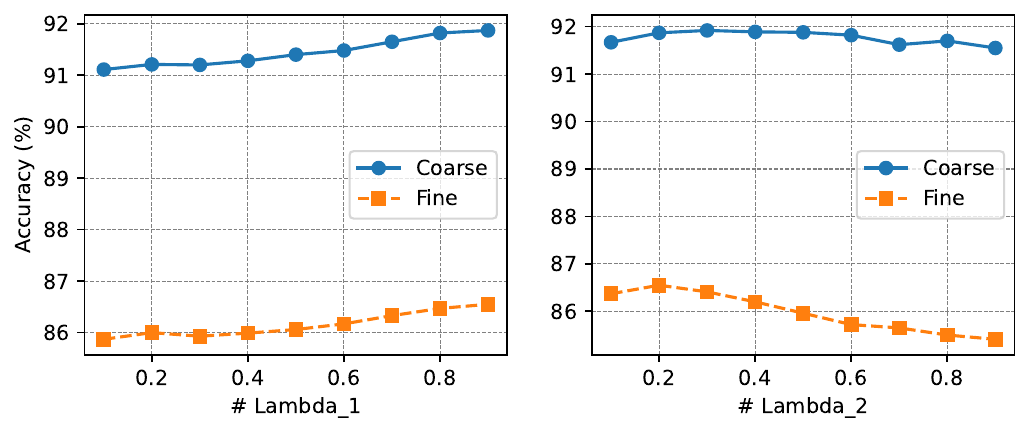}
    \caption{\centering Ablation on $\lambda_1$ and $\lambda_2$ on CIFAR-100 across coarse and fine level.}
    \label{fig:lambda}
    % \vspace{-1.5em}
\end{figure}
\subsection{Hyperparameter Settings} 
\label{sup:c1}
In Figure~\ref{fig:lambda}, We conduct ablation on $\lambda_1$ and $\lambda_2$ hyper-parameters of logits with structured features for CIFAR-100 dataset. Overall, an increase in $\lambda_1$ can enhance performance across multiple hierarchical levels, indicating that the textual structured information is rich and effective. On the other hand, with variations in $\lambda_2$, the performance across various levels peaks around $\lambda_2$ = 0.2. This reflects that structured image features provide more effective information, which in turn improves model performance. \\ 
\subsection{Analysis of Noisy Hierarchies Caused by LLMs} 
\label{sup:c2}
Taking Food-101 as an example, we conduct two types of comparative experiments utilizing hierarchical labels obtained from the LLMs (\textit{i.e.,} ChatGPT~\cite{openai2023chat}) (in Table~\ref{tab:llm}). (\textit{i}) The first experiment involves repeating the same input and testing the various hierarchical labels generated by the LLMs. Despite LLMs generating suboptimal hierarchical labels, our method still enhances model performance based on these noisy hierarchical labels. This demonstrates the strong robustness and generalizability of \model. (\textit{ii}) The second experiment involves generating hierarchical labels of different depths and conducting experiments based on these hierarchies. Across varying numbers of levels in label hierarchies, \model\ improves model performance, even though some categories in the label hierarchies may be overly broad or detailed. These experiments involve the performance disparity between GPT-mapped hierarchical labels and the officially provided annotation hierarchy, including inconsistencies in the labels or hierarchical levels. 
Through the validation of these experiments, our approach effectively utilizes the label hierarchy structure, thus improving hierarchical classification performance.
    \begin{table}[htbp]
        \centering
        \footnotesize
        \resizebox{\linewidth}{!}{
        \begin{tabular}{llccc}
            \toprule
            \multirow{2}{*}{Type} & \multirow{2}{*}{Method}  & \multicolumn{3}{c}{Acc. \%} \\ \cmidrule(r){3-3} \cmidrule(r){4-4} \cmidrule(r){5-5} &  & $l_1$  & $l_2$ & $l_3$   \\
            \midrule
            \multirow{3}{*}{Ground-Truth Hierarchical Labels} & CLIP & 61.38 & 85.53 & - \\ 
            & MaPLe & 88.16 & 88.04 & - \\ 
            & \model\ & \textbf{91.12} & \textbf{88.73} & - \\ \hdashline[1pt/1pt]
            \multirow{3}{*}{GPT-mapped 2-level Hierarchical Labels} & CLIP & 59.22 & 85.53 & -  \\
            & MaPLe & 87.55 & 87.26 & - \\ 
            & \model\  & 90.67 & 87.20 & - \\  \hdashline[1pt/1pt]
            \multirow{3}{*}{GPT-mapped 3-level Hierarchical Labels} & CLIP & 29.49 & 54.54 & 85.53 \\
            & MaPLe & 87.14 & 90.62 & 86.30 \\ 
            & \model\ & 90.06 & 93.35 & 88.56 \\ \bottomrule 
        \end{tabular}
        }
        \caption{Results based on GPT-mapped and annotated hierarchical labels.}
        \label{tab:llm}
    \end{table}

\section{Results Presentation}
\label{sup:d}
Our comparison methods are primarily divided into visual-only fine-grained visual classification methods and CLIP-based fine-tuning methods. The former is a uni-modal approach, utilizing only visual features and implementing hierarchical classification through multiple classification layers. We implement state-of-the-art methods from the past three years for comparison \cite{du2020fine,chang2021your,zhao2021graph,liu2022focus,xu2023trusted}. The latter comprises CLIP-based multi-modal methods, categorized into three types: zero-shot, feature adaptation, and prompt tuning. For zero-shot, we consider the only two methods \cite{ge2023improving,novack2023chils} combining class hierarchy and CLIP available to date in Appendix \ref{sup:d2}. We implement several popular methods of feature adaptation \cite{gao2023clip,zhang2022tip,guo2023calip,song2024meta}. Prompt tuning is the most common method for fine-tuning CLIP. We compare several popular methods from the past three years \cite{zhou2022learning,zhou2022conditional,jia2022visual,yao2023visual,khattak2023maple,khattak2023self}. 
% However, due to space constraints, we cannot present all methods' results on all datasets in the main text. 
% Therefore, we will provide detailed results in the Appendix \ref{sup:d5} and \ref{sup:d6}.
\\ Furthermore, it is important to note for experimental fairness that we are the first to migrate CLIP to hierarchical image classification, so there are no directly comparable methods that utilize both class hierarchy and CLIP. Currently, there are only two zero-shot methods that simultaneously utilize CLIP and class hierarchy. They are training-free, but we also compare them.

\begin{figure}[htbp]
    \centering
    \includegraphics[width=0.8\linewidth]{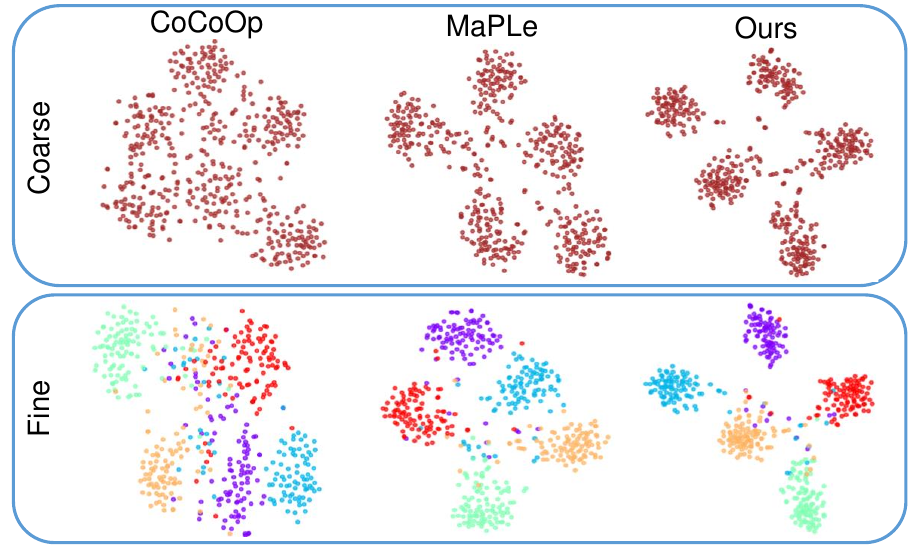}
    \caption{\centering Learned embedding tSNE visualization.}
    \label{fig:tsne1}
\end{figure}
\subsection{t-SNE visualization}
\label{sup:d1}
In Figure~\ref{fig:framework}, we present the visualization of image embeddings, which demonstrates that our method effectively separates categories across multiple granularities. However, it is hard to observe whether the features of certain categories with hierarchical relationships are separable. To intuitively understand the role played by the graph representations leveraging class hierarchy, we visualize the features of one single coarse category and its corresponding five fine-grained categories randomly selected from CIFAR-100 in Figure~\ref{fig:tsne1}. It is evident that the features of the fine-grained categories from \model\ are more distinct compared to both CLIP and MaPLe~\cite{khattak2023maple}. This can be attributed to the utilization of graph representations that integrate hierarchical relationships between categories. Similarly, the features at the coarse-grained level also benefit. CLIP does not focus on the intra-class arrangement of samples within the feature space, as long as these samples are proximate to each other and are separable from samples of different categories. This can lead to a chaotic distribution of subcategories within the same category region. In contrast, \model\ not only retains the attribute features of the coarse categories but also reduces the intra-class variance, achieving enhanced generalization. \\  %\ref{fig:tsne}
\begin{table}[htbp]
    \centering
    \footnotesize
    \resizebox{\linewidth}{!}{
    \begin{tabular}{lcccc}
    \toprule
    \multirow{2}{*}{Method}  & \multicolumn{2}{c}{CIFAR-100} & \multicolumn{2}{c}{Food-101}\\ \cmidrule(r){2-3}  \cmidrule(r){4-5}  & $l_1$  & $l_2$ & $l_1$  & $l_2$     \\
    \midrule
    CLIP~\cite{radford2021learning} & 43.22      &  66.57  & 61.38  &  85.53  \\
    CHiLS~\cite{novack2023chils}  & 52.58 & - & -  &  83.18 \\
    Hierarchy-CLIP~\cite{ge2023improving}  & 58.61  & 67.39  & 63.74 &  86.98  \\ \rowcolor{gray!20}
    \model~\textcolor{gray}{\tiny \textup{(Ours)}} & \textbf{91.87} & \textbf{86.55} & \textbf{91.12}   &  \textbf{88.73} \\ \bottomrule
    \end{tabular}
    }
    \caption{Performance analysis with CHiLS and Hierarchy-CLIP.}
    \label{tab:chils}
    \vspace{-1.5em}
\end{table}
\subsection{Analysis with CHiLS~\cite{novack2023chils} and Hierarchy-CLIP~\cite{ge2023improving}} 
\label{sup:d2}
These two works are among the most recent and earliest to combine label hierarchies with VLMs. Specifically, CHiLS utilizes the subcategories of each label to enrich the class name itself, which is effective in cases where datasets may contain rich underlying structures but have information-poor class labels. However, this approach experiences a notable performance decline when applied to some fine-grained datasets. This is because the class labels in fine-grained datasets are already very specific, and forcibly adding more specialized and specific subcategories can overwhelm the model, leading to an inability to match images with the correct text. Building upon this, Hierarchy-CLIP enriches class labels to a greater extent by adding both the superclass and subclasses to the class name after initially identifying the top-5 class candidates. This effectively compensates for the shortcomings of CHiLS. Nevertheless, this method is still training-free and cannot be maximally adapted to downstream datasets. Fine-tuning methods based on CLIP, whether prompt tuning~\cite{zhou2022conditional,zhou2022learning,khattak2023maple,jia2022visual} or feature adaptation~\cite{gao2023clip,zhang2022tip,guo2023calip}, do not integrate label hierarchies. Therefore, the introduction of \model\ effectively addresses these limitations. It is trainable and can be easily transferred to datasets of different semantic granularities (regardless of the availability of label hierarchies), and it also enhances the classification accuracy of each semantic hierarchical category in the dataset. In Table~\ref{tab:chils}, we compare the performance of our method with that of these two approaches. Regarding the CHiLS experiment, as their paper indicates that it is tested only on the 20 classes of CIFAR-20 and 101 classes of Food-101, to maximize the representation of its effect, we follow its setup and do not test it on the remaining hierarchical categories. The results show an improvement on the generic image dataset, but a decline on the fine-grained food dataset, consistent with our previous analysis. In the Hierarchy-CLIP experiment, it also exhibits commendable performance, showing improvements across various hierarchical levels. This is largely attributed to its use of both superclasses and subclasses of labels for expansion. Our method not only enhances classification performance at each level but also can be conveniently transferred to downstream datasets.\\ 
\subsection{Params number and FLOPs}
\label{sup:d3}
Table \ref{tab:para} shows the computational complexity of \model\ in comparison with previous works. Although \model\ leverages graph encoders, its overall FLOPs are only 0.1\% higher than MaPLe, and it offers better convergence and competitive training/inference speed.\\
\begin{table}[htbp]
    \centering
    \scriptsize
    \resizebox{\linewidth}{!}{
    \begin{tabular}{llcccccccc}
    \toprule
    Type &  Method  & Params & \begin{tabular}[c]{@{}c@{}}Params\\\%CLIP\end{tabular} & GFLOP & \begin{tabular}[c]{@{}c@{}}Train time\\(min)\end{tabular}  & FPS & Acc. \% \\ 
    \midrule
    \multirow{4}{*}{Prompt Learning} & CoOp~\cite{zhou2022learning} & 2k & 0.002 & 163.4 & 13.52 & 1345 & 80.29 \\
    & CoCoOp~\cite{zhou2022conditional}  & 36k & 0.03 & 163.4 & 42.18 & 15.3 & 79.17  \\
    & MaPLe~\cite{khattak2023maple} & 3.55M & 2.85 & 192.7 & 22.34 & 1357  & 88.24 \\
    & PromptSRC~\cite{khattak2023self} & 32k & 0.02 & 181.1 & 15.43 & 1379  & 85.21 \\ \hdashline[1pt/1pt]
    \multirow{2}{*}{Feature Adaptation} & CLIP-Adapter~\cite{gao2023clip} & 0.52M & 0.42 & 178.2 & 54.32 & 1298 & 80.47 \\ 
    & Tip-Adapter~\cite{zhang2022tip} & 0.4k & 3e-6 & 203.0 & 7.16 & 1466 & 83.00 \\  \hdashline[1pt/1pt]
    \multirow{4}{*}{Visual-only FGVC} & PMG~\cite{du2020fine} & 27.8M & - & 123.05 & 69.20 & 42.17 & 85.09 \\
    & FGN~\cite{chang2021your} & 29.3M & - & 120.17 & 78.57 & 38.04 & 85.74 \\
    & CHRF~\cite{liu2022focus} & 31.6M & - & 118.42 & 89.15 & 33.50 & 86.79 \\ 
    & TFGIC~\cite{xu2023trusted} & 31.8M & - & 119.56 & 88.06 & 36.36 & 87.19 \\ \hdashline[1pt/1pt]
    \rowcolor{gray!20}
    & \model$\dagger$ & 2.15M & 1.73 & 191.3 & 18.44 & 1349 & \underline{88.82} \\ \rowcolor{gray!20}
    & \model & 4.25M & 3.41 & 193.6 & 20.17 & 1349  & \textbf{89.21} \\ \bottomrule
    \end{tabular}}
    \caption{Computing cost comparison on CIFAR-100 dataset. $\dagger$ denotes the GNN parameters of the two modalities are shared. FPS denotes Frames Per Second (with a batch size of 100).}
    \label{tab:para}
    \vspace{-1.5em}
\end{table}

\end{document}